\documentclass[runningheads]{llncs}
\usepackage{wrapfig}
\pdfoutput=1
 
\usepackage{eccv}
\usepackage[ruled]{algorithm2e}
\usepackage[ruled]{algorithm2e}
\usepackage{xcolor}
\definecolor{darkgreen}{rgb}{0.0, 0.5, 0.0}

\usepackage{eccvabbrv}

\usepackage{graphicx}
\usepackage{booktabs}

\usepackage[accsupp]{axessibility}  

\usepackage{hyperref}

\usepackage{orcidlink}

\begin{document}


\title{Fine-Grained Scene Graph Generation via Sample-Level Bias Prediction} 

\titlerunning{Fine-Grained Scene Graph Generation via Sample-Level Bias Prediction}

\author{Yansheng Li\inst{1} \and Tingzhu Wang\inst{1}$^\dag$ \and
Kang Wu\inst{1} \and Linlin Wang\inst{1}   \and \\  Xin Guo\inst{2} \and Wenbin Wang\inst{3} }

\authorrunning{Y.~Li et al.}

\institute{School of Remote Sensing and Information Engineering, Wuhan University \and Ant Group \and Hubei Key Laboratory of Intelligent Vision Monitoring for Hydroelectric Engineering, College of Computer and Information Technology, China Three Gorges University\\
\email{tingzhu.wang@whu.edu.cn}}
\maketitle

\begin{abstract}
Scene Graph Generation (SGG) aims to explore the relationships between objects in images and obtain scene summary graphs, thereby better serving downstream tasks. However, the long-tailed problem has adversely affected the scene graph's quality. The predictions are dominated by coarse-grained relationships, lacking more informative fine-grained ones. The union region of one object pair (i.e., one sample) contains rich and dedicated contextual information, enabling the prediction of the sample-specific bias for refining the original relationship prediction. Therefore, we propose a novel Sample-Level Bias Prediction (SBP) method for fine-grained SGG (SBG). Firstly, we train a classic SGG model and construct a correction bias set by calculating the margin between the ground truth label and the predicted label with one classic SGG model. Then, we devise a Bias-Oriented Generative Adversarial Network (BGAN) that learns to predict the constructed correction biases, which can be utilized to correct the original predictions from coarse-grained relationships to fine-grained ones. The extensive experimental results on VG, GQA, and VG-1800 datasets demonstrate that our SBG outperforms the state-of-the-art methods in terms of Average@K across three mainstream SGG models: Motif, VCtree, and Transformer. Compared to dataset-level correction methods on VG, SBG shows a significant average improvement of 5.6\%, 3.9\%, and 3.2\% on Average@K for tasks PredCls, SGCls, and SGDet, respectively. The code will be available at \url{https://github.com/Zhuzi24/SBG}.

  \keywords{Scene graph generation \and Long-tailed distribution \and Bias correction \and Fine-grained relationships}
\end{abstract}


\renewcommand{\thefootnote}{\fnsymbol{footnote}}
\footnotetext[0]{$^\dag$ indicates the corresponding author.}

\section{Introduction}
\label{sec:intro}

 In recent years, Scene Graph Generation (SGG) has emerged as a popular area of research. SGG aims to generate a structured summary graph from an image. This graph consists of nodes and edges, where nodes represent objects in the image, with each node containing category and location information. The edges represent the relationships between the objects. The structured semantic graph clearly describes the objects, their attributes, and the relationships between them in the scene. It offers the abundant semantic information and has the powerful representational capability. Therefore, it can strongly support various downstream tasks, including visual question answering~\cite{VQA2,VQA3}, image retrieval~\cite{IR1,IR2}, and image captioning~\cite{IC1,IC2}.

\begin{figure}[tb]
  \centering
  \includegraphics[width=0.75\linewidth]{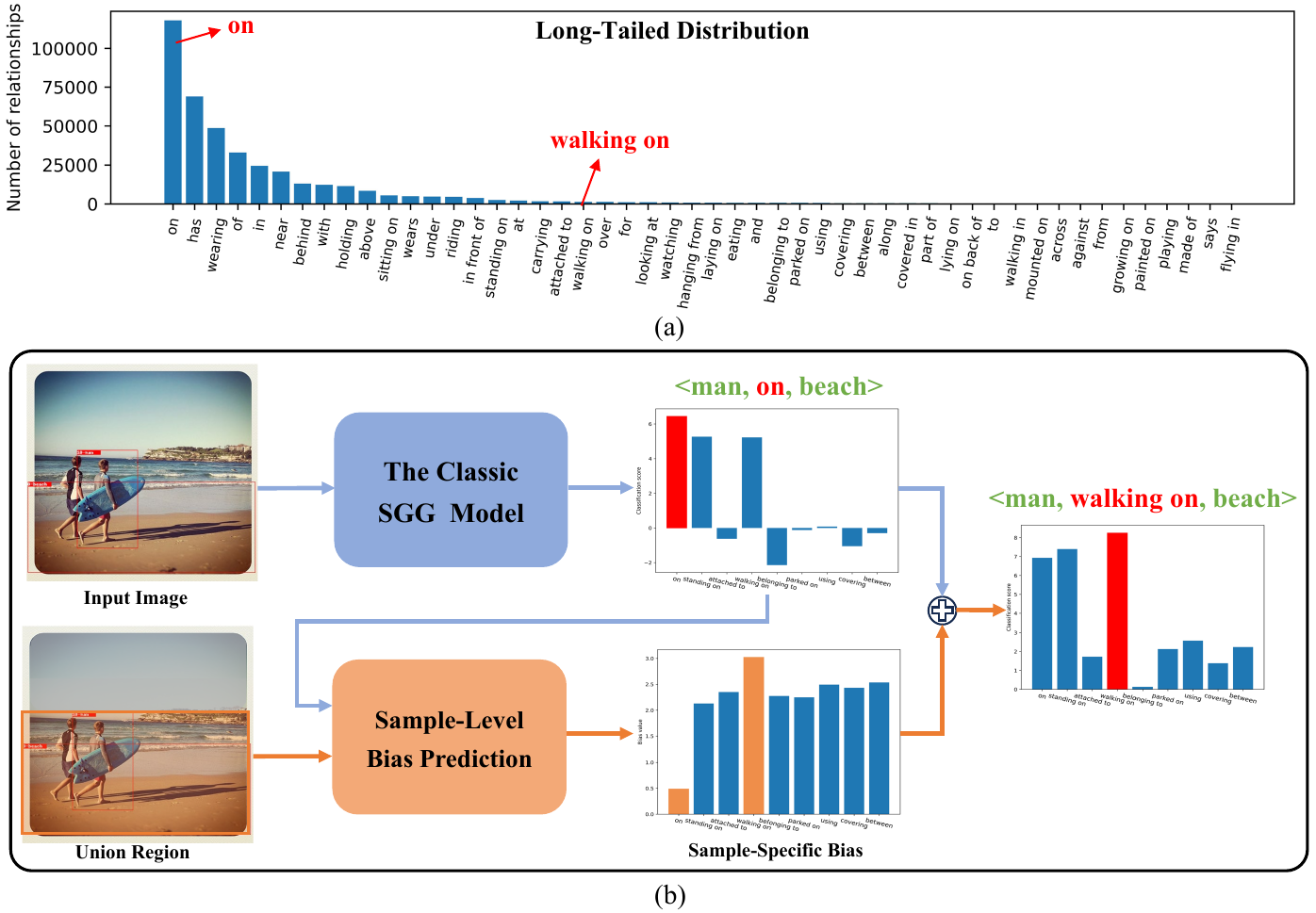}

  \caption{(a) The long-tailed distribution of relationships in the well-known Visual
Genome (VG) dataset~\cite{vg}. (b) Workflow for bias correction: a sample-specific correction bias is predicted by the contextual information from the union region of $<$man, beach$>$ and the original prediction, then it corrects the prediction from the coarse-grained ``\emph{on}'' to the fine-grained ``\emph{walking on}''. The y-axis label ``classification score'' in the figure represents the classification value before the softmax function.}

   \label{fig:fig_1}
\end{figure}

However, SGG suffers from the long-tailed problem, which is prevalent across most datasets. \cref{fig:fig_1}(a) depicts the long-tailed distribution of relationships in the well-known Visual Genome (VG) dataset~\cite{vg}.
There is a significant abundance of coarse-grained head categories, such as ``\emph{on}'', ``\emph{has}'', and ``\emph{wearing}'', whereas fine-grained tail categories, like ``\emph{walking on}'', ``\emph{part of}'', and ``\emph{flying in}'', are comparatively scarce. The long-tailed problem significantly impacts the model's predictions for relationships, as depicted in \cref{fig:fig_1}(b). The prediction of the classic SGG model for $<$man, beach$>$'s relationship is influenced by this issue. The model predicts a coarse-grained relationship ``\emph{on}'', which lacks the sufficient information. 

Several methods~\cite{T1, HS1, T3, HML2022, Pbiswas2023} have been proposed to mitigate the impact posed by the long-tailed problem. DLFE~\cite{HS1} and RTPB~\cite{T3} address this problem by correcting relationship predictions. DLEF employs the estimated per-class label frequency to recover the unbiased relationship probabilities from the biased ones. RTPB leverages the resistance bias to improve the model’s detecting ability for tail categories. Both of them share a common limitation in that they apply the estimated per-class label frequency and resistance bias for all object pairs' relationship predictions, neglecting the specificity of each object pair (i.e., each sample). They can be categorized as the dataset-level correction, as the estimated per-class label frequency and resistance bias are the overall information of the dataset. In contrast to them, we are dedicated to focusing on capturing the characteristics of each object pair. The contextual information differs among different object pairs' union regions, enabling us to predict sample-specific biases for correcting the corresponding relationship predictions. This is also denoted as the sample-level bias correction. The prior information of relationships typically facilitates the generation of the scene graph, so we incorporate relationships' global bias into the bias prediction process. The global bias is the relationship's prior bias obtained through the statistics of the dataset. The process of bias correction is depicted in \cref{fig:fig_1}(b), a sample-specific bias is predicted to correct the coarse-grained ``\emph{on}'' to the fine-grained ``\emph{walking on}''.

Inspired by the above idea, we propose a novel Sample-Level Bias Prediction (SBP) method for fine-grained SGG (SBG). We first train a classic SGG model and construct a correction bias set by calculating the margin between the ground truth label and the predicted label with one classic SGG model. Each correction bias in the set represents a vector that contains bias values for each relationship category and is designed to accurately correct the corresponding original prediction. Next, we aim to predict a vector that closely approximates the constructed correction bias. The conventional methods, such as fully connected networks~\cite{schwing2015fully} and one-dimensional convolutional networks~\cite{malek2018one}, have relatively simple loss constraints and weak generative capabilities. Therefore, we devise a Bias-Oriented Generative Adversarial Network (BGAN). BGAN employs an adversarial training approach where the generator and discriminator compete with each other. This adversarial learning process empowers the BGAN with robust generative capabilities to generate the desired vector. The constructed correction bias set is utilized to train the generator and discriminator in BGAN. Finally, the generator in BGAN predicts sample-specific bias for refining the original predictions from the classic SGG model, further obtaining fine-grained relationships.

To assess the effectiveness of our SBG, we introduce the Average@K~\cite{HML2022, CFAl}, which can better reflect the overall performance of all relationships. In sum, this paper provides the following valuable contributions to the SGG community in the hope of advancing the
fine-grained SGG research:

\begin{itemize}
    \item This paper, for the first time, explores sample-level bias correction to tackle the long-tailed problem, promoting refining the coarse-grained relationships into the fine-grained ones.
    \item We devise a novel Bias-Oriented Generative Adversarial Network (BGAN), which leverages the contextual information to predict the sample-specific correction bias.
    \item The experiments on VG, GQA, and VG-1800 datasets demonstrate that our SBG outperforms the state-of-the-art methods and shows a significant improvement compared to dataset-level correction methods on Average@K.

\end{itemize}

\section{Related Work}
\subsection{Scene Graph Generation}
SGG~\cite{li2021interventional, lu2016visual, xie2017aggregated, pcpl} decodes images into structured semantic graphs, supporting various downstream tasks~\cite{VQA2,IC1,IR1,IR2,luo2024skysensegpt}. Early works were primarily dedicated to designing models to achieve the relationship predictions between the objects. \cite{2016vis} leveraged language priors from semantic word embeddings to finetune the predicted relationships, \cite{2017Drn} exploited the statistical dependencies between objects and their relationships, and \cite{2018rank} designed a ranking objective function by enforcing the annotated relationships to have higher relevance scores. The method described above neglected the rich visual context information from images. To address this problem, some methods designed elaborate feature refinement modules to encode the visual context information, such as message passing strategies~\cite{BGNN, IMP}, sequential LSTMs~\cite{Motif, Vctree}, graph neural networks~\cite{chen2019knowledge, zareian2020bridging, lin2020gps}, and self-attention networks~\cite{sharifzadeh2021classification, lu2021context, gkanatsios2019attention}. In addition, there are now some datasets as well as methods~\cite{li2024scene,li2021semantic} emerging in the remote sensing field.

Later works focused on designing optimization frameworks to further improve the performance of SGG models. Many optimization frameworks were proposed to tackle the long-tailed problem of relationships. \cite{T1} used causal inference for unbiased SGG, \cite{EMB} proposed an energy-based constraint loss to learn relationships in small numbers. \cite{T2,HML2022} employed hierarchical or grouped strategies to gradually learn, thereby mitigating the impact of the long-tailed effect. \cite{HS1,T3} corrected the relationship predictions to obtain the unbiased scene graph. \cite{Pbiswas2023} leveraged a within-triplet Bayesian network to eradicate the long-tailed effect. ~\cite{CFAl} proposed a compositional feature augmentation strategy to mitigate the bias from the perspective of increasing the diversity of triplet features. There were also some methods using data augmentation strategies to improve the performance of SGG models, such as \cite{guo2021general,desai2021learning,IE}. Optimization frameworks offer greater flexibility and improved malleability, so it is a promising research direction.

\begin{figure}[!]
    \begin{center}
        \includegraphics[width=0.9\linewidth]{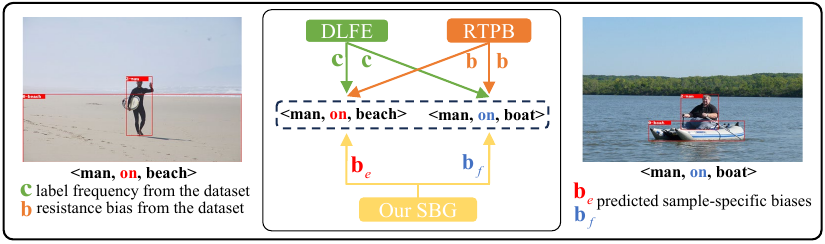}
    \end{center}
   
    \caption{Compare our SBG with DLFE~\cite{HS1} and RTPB~\cite{T3} for corrections. DLFE and RTPB apply the same $\mathbf{c}$ or $\mathbf{b}$ to correct all predictions while our SBG predicts the sample-specific bias for each prediction.
    }
    \label{fig:fig_com_D_R_Our}
\end{figure}

\subsection{Long-Tailed Learning}
Recently, several correction methods have emerged to address the long-tailed problem in SGG. In \cref{fig:fig_com_D_R_Our}, DLFE~\cite{HS1} and RTPB~\cite{T3} employed different ways to correct the original predictions. DLFE estimated the per-class label frequency $\mathbf{c}$ and used it to correct the biased probabilities. By dividing the biased probabilities by $\mathbf{c}$, DLFE obtained unbiased probabilities. RTPB utilized the resistance bias $\mathbf{b}$ to enhance the model's focus towards tail relationships. RTPB adjusted the relationship’s classification logits by subtracting $\mathbf{b}$, thereby correcting the biased relationship predictions.

In DLFE, $\mathbf{c}$, and in RTPB, $\mathbf{b}$, are both vectors that capture the overall information of relationships in the dataset. As shown in \cref{fig:fig_com_D_R_Our},  DLFE and RTPB apply the same $\mathbf{c}$ or $\mathbf{b}$ to correct all predictions, so they do not sufficiently consider the specificities of different object pairs. They are the sample-insensitive bias correction which is also denoted as the dataset-level bias correction. In contrast, our SBG focuses on the characteristics of each object pair to predict the sample-specific bias for refining the original relationship predictions. Our SBG is the sample-sensitive bias correction which is also denoted as the sample-level bias correction.

\section{Method}

\subsection{Problem Setting \& Overview Framework}
\noindent\textbf{Problem Setting.} The task of SGG is to generate a connected summary graph $\mathcal{G}$ from an input image $\textit{I}$. The common SGG methods are the two-stage processes involving object detection followed by relationship prediction. In the first stage, we detect all objects in the image $\textit{I}$, denotes as $\mathcal{O}=\{o_{i}\}$. In the second stage, we predict relationships $\mathcal{R}=\{r_{i\rightarrow j}\}$ of object pairs ($o_{i}$, $o_{j}$) and then generate ($o_{i}$,$r_{i\rightarrow j}$,$o_{j}$) triplets. Therefore, the scene graph can be formulated as:
\begin{equation}
\mathcal{G} = \{(o_i,r_{i\rightarrow j},o_j)|o_i, o_j\in\mathcal{O}, r_{i\rightarrow j}\in \mathcal{R}\}.
\end{equation}

\begin{figure*}[t]
    \begin{center}
        \includegraphics[width=1\linewidth]{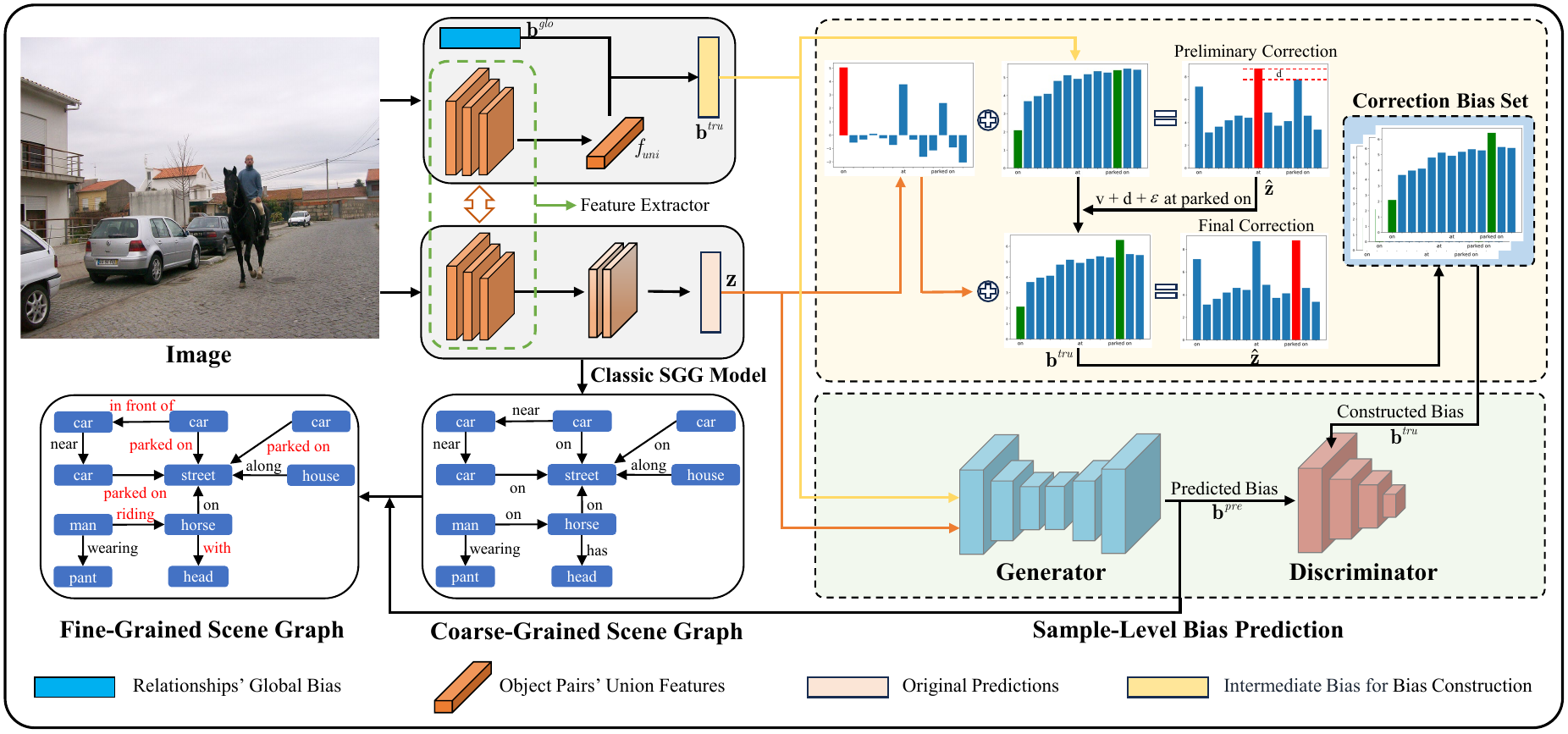}
    \end{center}
   
    \caption{The overall structure of our SBG. After constructing the correction bias set, BGAN learns to predict the constructed correction biases for achieving the sample-level bias correction. The coarse-grained scene graph is generated by the classic SGG model. v is the value at ``parked on'' in $\mathbf{b}^{tru}$. }
    \label{fig:overall}
    
\end{figure*}

\noindent\textbf{Overall Framework.} \cref{fig:overall} illustrates the overall structure of SBG, and the entire workflow follows the common two-stage SGG pipeline. Consistent with previous methods~\cite{T1,T3}, we utilize Faster R-CNN~\cite{ren2015faster} as the object detection network. To address the long-tailed problem of relationships, we present a novel method SBP to realize the sample-level bias correction. More detailed analysis can be seen in \cref{sec:WGAN}.

\subsection{Sample-Level Bias Prediction}
\label{sec:WGAN}
The unbiased relationship predictions can be recovered from the biased ones~\cite{HS1}. In an ideal scenario, training the SGG model on an unbiased dataset results in unbiased relationship predictions which are denoted as $\mathbf{z}_{u}=[z_{u1},z_{u2},...,z_{uM}]$, where $M$ is the number of relationship categories. However, datasets often suffer from the severe long-tailed problem of relationships, thereby leading to biased relationship predictions which are represented by $\mathbf{z}=[z_{1},z_{2},...,z_{M}]$. Both $\mathbf{z}_{u}$ and $\mathbf{z}$ are vectors of logits for relationship classification. If we can obtain the differences which are denoted as $\mathbf{b_{u}}=[b_{u1},b_{u2},...,b_{uM}]$ between $\mathbf{z}_{u}$ and $\mathbf{z}$, where an ideal $\mathbf{b_{u}}$ equals $\mathbf{z}_{u}-\mathbf{z}$, then we can recover the unbiased predictions $\mathbf{z}_{u}$ from the biased predictions $\mathbf{z}$. However, in reality, obtaining $\mathbf{b_{u}}$ is challenging as acquiring $\mathbf{z}_{u}$ is nearly impossible. This poses difficulties in recovering $\mathbf{z}_{u}$ from $\mathbf{z}$.

An alternative perspective can provide a further insight. $\mathbf{z}_{u}$ can accurately predict relationships while other data vectors which are denoted as $\hat{\mathbf{z}}$ may also achieve accurate relationship predictions. Although $\mathbf{z}_{u}$ and $\hat{\mathbf{z}}$ may have significant differences, their predicted relationships are the same. The similarity between $\mathbf{z}_{u}$ and $\hat{\mathbf{z}}$ here can be expressed as $argmax(softmax(\mathbf{z}_{u}))=argmax(softmax(\hat{\mathbf{z}}))$, and the difference between them is that $\mathbf{z}_{u}$ is ideal and unbiased and $\hat{\mathbf{z}}$ is realizable and not unbiased. In reality, obtaining $\hat{\mathbf{z}}$ is relatively easier compared to acquiring $\mathbf{z}_{u}$. Images contain abundant information, offering possibilities for generating $\hat{\mathbf{z}}$. The union regions of object pairs contain rich and dedicated contextual information, enabling us to predict the sample-specific biases $\mathbf{b_s}$, ensuring $\hat{\mathbf{z}}=\mathbf{z}+\mathbf{b_s}$. By applying the sample-specific biases $\mathbf{b_s}$ to $\mathbf{z}$, we achieve the sample-level bias correction for the long-tailed problem.

To realize the aforementioned idea, we propose the SBP to predict the sample-specific bias $\mathbf{b_s}_{i,j}$ for correcting the biased relationship prediction $\mathbf{z}_{i,j}$, where $\mathbf{z}_{i,j}$ denotes the relationship prediction of the $j$-th object pair in the $i$-th image. The parameters of the classic SGG model and SBP are denoted as $\Lambda^{O}$ and $\Lambda^{B}$, respectively. Let $\mathbf{z}_{i,j}=\Upsilon(\Lambda^{O};I_{i})$ and $\mathbf{b_s}_{i,j}=\Upsilon(\Lambda^{B};f_{uni}^{i,j},\mathbf{b}^{glo},\mathbf{z}_{i,j})$ represent the outputs of the classic SGG model and SBP, respectively. Thus, the sample-level bias correction for the original prediction can be expressed as:
\begin{equation}
\begin{aligned}
\hat{\mathbf{z}_{i,j}}&=\Upsilon(\Lambda^{O};I_{i})+\Upsilon(\Lambda^{B};f_{uni}^{i,j},\mathbf{b}^{glo},\mathbf{z}_{i,j})\\
               &=\mathbf{z}_{i,j}+\mathbf{b_s}_{i,j},
\label{eq: bias}
\end{aligned}
\end{equation}
where $\Upsilon(\cdot; \cdot)$ represents the process of obtaining the neural network output based on the neural network's parameters and input data. $f_{uni}^{i,j}$ is the union feature of the $j$-th object pair in the $i$-th image. It can be obtained by extracting the visual feature of the object pair's union region using a shared weight feature extractor with Faster R-CNN in the classic SGG model (as shown in \cref{fig:overall}), and then fusing it with the spatial feature of object pair's union region. $\mathbf{b}^{glo}$ is the relationships' global bias, serving as the prior bias of the relationships. It is calculated as $-log(w^{a}/ {\textstyle \sum_{j \in M}^{}} w_j^a+ \epsilon)$~\cite{T3}, where $w$ represents the weights of the relationships, and $a$ and $\epsilon$ are hyper-parameters. $\hat{\mathbf{z}_{i,j}}$ is relationship prediction after the sample-level bias correction. To guide the training of SBP, we construct a set of correction biases $\mathbf{b}^{tru}$ where the set is denoted as $\mathcal{S}=\{\mathbf{b}^{tru}\}$ according to the ground truth labels by \cref{eq: bias}.

\begin{figure}[ht]

\centering 
\hspace*{0.1cm}
\begin{minipage}{0.425\textwidth}  
    \centering  
    \caption{The construction workflow of Correction Bias Set $\mathcal{S}$.}
    \includegraphics[width=\linewidth]{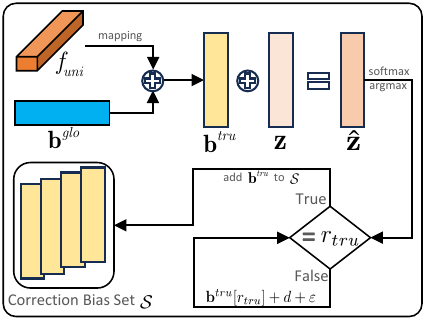}
    
    \label{fig:CBS}
\end{minipage}
\begin{minipage}{0.55\textwidth} 
\centering  
\hspace*{0.7cm}
\resizebox{0.75\linewidth}{!}{  
    \begin{algorithm}[H]
        \caption{The training of BGAN for one iteration.}
        \label{SBP}
        \LinesNumbered
        \KwIn{$f_{uni}$, $\mathbf{b}^{glo}$, $\mathbf{z}$, and $\mathbf{b}^{tru}$}
    
        \For{$k\leftarrow 1$ \KwTo $5$}{
            1. clear the grad of $D$\;
            2. $\mathbf{b}^{pre} = \Upsilon(\Lambda^{B_{G}};f_{uni},\mathbf{b}^{glo},\mathbf{z})$\;
            3. calculate $\mathcal{L}_{D}$ using $\mathbf{b}^{pre}$ and $\mathbf{b}^{tru}$\;
            4. update the parameters $\Lambda^{B_{D}}$ of $D$\;
        \If{$k=5$}{
            1. clear the grad of $G$\;
            2. $\mathbf{b}^{pre} = \Upsilon(\Lambda^{B_{G}};f_{uni},\mathbf{b}^{glo},\mathbf{z})$\;
            3. calculate $\mathcal{L}_{G}$ using $\mathbf{b}^{pre}$\;
            4. update the parameters $\Lambda^{B_{G}}$ of $G$\;
            }  
        }
        update the learning rates of $D$ and $G$.
    \end{algorithm}
 }

\end{minipage}
\end{figure}

\noindent\textbf{Construction of Correction Bias Set.} In each training iteration, $\mathcal{S}$ is constructed
by calculating the margin between the ground truth label $r_{tru}$ and the predicted label $r_{pre}$ from one classic SGG model. The construction workflow of $\mathcal{S}$ is summarized in \cref{fig:CBS}. The fusion of ${f_{uni}}$ and $\mathbf{b}^{glo}$ results in $\mathbf{b}^{tru}=\phi(f_{uni})+\mathbf{b}^{glo}$, where $\phi$ denotes the mapping of high-dimensional feature to one-dimensional feature using an encoder that includes a single layer of transformer~\cite{vaswani2017attention}. Subsequently, we assess whether $\mathbf{b}^{tru}$ meets the condition $r_{tru}=r_{pre}$, where $r_{pre}=argmax(softmax(\hat{\mathbf{z}}))$ and $\hat{\mathbf{z}}=\mathbf{z}+\mathbf{b}^{tru}$. If $\mathbf{b}^{tru}$ satisfies the condition, it is included in $\mathcal{S}$; otherwise, the difference $d=\hat{\mathbf{z}}[r_{tru}]-\hat{\mathbf{z}}[r_{pre}]$ is calculated. After updating $\mathbf{b}^{tru}[r_{pre}]=\mathbf{b}^{tru}[r_{pre}]+d+\varepsilon$, the modified $\mathbf{b}^{tru}$ is added to $\mathcal{S}$, where $\varepsilon$ denotes a small non-zero value.

\noindent\textbf{Training of Bias Prediction.}
The constructed $\mathcal{S}$ can accurately correct the original predictions. We aim to learn to predict the constructed biases. The closer the predicted biases are to the constructed biases, the better the effectiveness of the bias correction. To achieve this target, we propose the BGAN, which employs one-dimensional convolutions to model the prediction process. The generator (denoted as $G$) in BGAN takes the object pairs' union features $f_{uni}$, the relationships' global bias $\mathbf{b}^{glo}$, and the original predictions $\mathbf{z}$ as input to predict the sample-specific biases $\mathbf{b}^{pre}$. The predicted biases $\mathbf{b}^{pre}$ and the constructed biases $\mathbf{b}^{tru}$ from $\mathcal{S}$ are fed into the discriminator (denoted as $D$) in BGAN for the training. Through the adversarial training between $G$ and $D$, $G$ learns to predict the sample-specific biases for correcting the original predictions. Additionally, the cross-entropy loss~\cite{CEloss} is computed using the corrected predictions $\hat{\mathbf{z}}$ to further constrain the generator $G$. The inputs and outputs of $G$ and $D$ can be represented as:
\begin{equation}
\mathbf{b}^{pre} = \Upsilon(\Lambda^{B_{G}};f_{uni},\mathbf{b}^{glo},\mathbf{z}),
\end{equation}
\begin{equation}
\mathcal{T}_{G} = \Upsilon(\Lambda^{B_{D}};\mathbf{b}^{pre}), \mathcal{T}_{S} = \Upsilon(\Lambda^{B_{D}};\mathbf{b}^{tru}),
\end{equation}
where $\Lambda^{B_{G}}$ and $\Lambda^{B_{D}}$ represent the parameters of $G$ and $D$, respectively, and they both belong to the parameters $\Lambda^{B}$. The $\mathcal{T}_{G}$ and $\mathcal{T}_{S}$ represent the discriminative scores of $D$ for $\mathbf{b}^{pre}$ and $\mathbf{b}^{tru}$.
The losses of $G$ and $D$ can be expressed as:
\begin{equation}
\mathcal{L}_{G} = -mean(\mathcal{T}_{G})+\alpha \cdot(-t_{m}log(\frac{e^{\mathbf{\hat{z}}_{m}}}{\sum_{m}^{M}e^{\mathbf{\hat{z}}_{m}}})),
\end{equation}
\begin{equation}
\mathcal{L}_{D} = -mean(\mathcal{T}_{S})+mean(\mathcal{T}_{G}),
\end{equation}
where $\alpha$ represents the weight factor for the cross-entropy loss. $t_{m}$ takes the value 1 only when $m$ is the ground truth label; otherwise, it is set to 0. The detailed training workflow of BGAN is outlined in \cref{SBP}. Thus, the overall loss is defined as follows:
\begin{equation}
\underset{\Lambda^{O},\Lambda^{B}}{\mathrm{min}} \mathcal{L}_{total}=\mathcal{L}_{SGG}+\beta \cdot \mathcal{L}_{BGAN},
\end{equation}
where $\mathcal{L}_{SGG}$ represents the cross-entropy loss of the classic SGG model, which takes $\mathbf{z}$ as input. $\mathcal{L}_{BGAN}$ stands for $\mathcal{L}_{G}$ and $\mathcal{L}_{D}$. During the training of the classic SGG model, $\beta$ is set to 0. After the training of the classic SGG model is completed, its parameters $\Lambda^{O}$ are frozen, and the training of BGAN is conducted by setting $\beta$ to 1. Once the training of BGAN is finished, the classic SGG model and the generator $G$ in BGAN are jointly used for inference. $G$ predicts the sample-specific biases to correct the original predictions, and further obtain the fine-grained relationships.

\section{Experiments}

\subsection{Experimental Settings}
\noindent\textbf{Datasets.} 
We conduct experiments on three datasets: VG, GQA, and VG-1800. In SGG, the most commonly used database is VG~\cite{vg}. From the original database, the most frequent 150 object and 50 relationship categories are retained~\cite{IMP,T1}. GQA~\cite{hudson2019gqa} is a vision-and-language dataset similar to VG, consisting of a total of 113k images. Following~\cite{Pbiswas2023}, we split the GQA dataset for the training. We retain images only with the most frequent 160 object and 60 relationship categories. VG-1800~\cite{IE} is a large-scale dataset, containing 70k object and 1.8k relationship categories. For both VG and GQA, we use 70$\%$ of the images for the training and the remaining 30$\%$ for the testing. Similar to \cite{T1}, we sample a 5k validation set for the parameter tuning. For VG-1800, settings we keep in line with IETrans~\cite{IE}.

\noindent\textbf{Tasks.} We perform three conventional SGG sub-tasks: (1) Predicate Classification (PredCls): given object bounding boxes and object labels, predicting the relationships between the objects. (2) Scene Graph Classification (SGCls): given object bounding boxes, predicting the object labels and relationships between the objects. (3) Scene Graph Detection (SGDet): performing object detection and then predicting relationships between the detected objects.

\noindent\textbf{Evaluation Metrics.} We evaluate SGG models on four metrics: (1) Recall@K (R@K)~\cite{Motif}: It indicates the proportion of ground truths that appear among the top-K confident predicted triplets. (2) mean Recall@K (mR@K)~\cite{T2}: It is the average of R@K scores which are calculated for each relationship category separately. (3) Average@K (A@K)~\cite{CFAl}: It calculates the average scores of R@K and mR@K. Since R@K favors head relationships while mR@K favors tail relationships, the A@K can provide a more intuitive and comprehensive reflection of the overall performance of all relationships. (4) F-Acc~\cite{IE}: It is a metric used for VG-1800 that indicates comprehensive performance of all relationships.

\subsection{Implementation Details}
\label{sec:v150}
\noindent\textbf{Object Detector.} For VG and VG-1800, following~\cite{HS1}, we employ a pre-trained Faster R-CNN~\cite{ren2015faster} with ResNeXt-101-FPN~\cite{xie2017aggregated} backbone. In the training stage, the parameters of the detector are fixed to reduce the computation cost. For GQA, we train a Faster-RCNN with ResNeXt-101-FPN as the object detector, and then the detector reaches 26.0 (\%) mAP on the test set.

\noindent\textbf{Model Settings.} For the classic SGG model, we set the batch size to 16 and employ a SGD optimizer~\cite{keskar2017improving} with an initial learning rate of 0.001. The training procedure contains a total of 18,000 iterations. For SBP, BGAN is comprised of one-dimensional convolutions. The number of layers for $G$ and $D$ are 5 and 3, respectively. Both of them employ an RMSProp optimizer~\cite{zou2019sufficient} with initial learning rates of 0.0001 and 0.0005. $\varepsilon$ is set to 0.0001. The training process contains a total of 4,000 iterations. All experiments are performed on NVIDIA 24G GeForce RTX 3090 GPUs.

\begin{table*}[!ht]
	\centering

    \small
     \setlength{\tabcolsep}{1mm}

   	\caption{Performance comparison with the state-of-the-art two-stage methods on PredCls, SGCls, and SGDet tasks of VG for R@50/100 (\%), mR@50/100 (\%), and A@50/100 (\%). The underlined values represent suboptimal performance. The light blue font belongs to the dataset-level correction.
	}
    \resizebox{\linewidth}{!}{
	\begin{tabular}{c|ccc|ccc|ccc}
\hline

 & \multicolumn{3}{c|}{PredCls}             & \multicolumn{3}{c|}{SGCls}                   & \multicolumn{3}{c}{SGDet}               \\ \hline
Model          & R@50/100    & mR@50/100   & A@50/100    & R@50/100      & mR@50/100     & A@50/100    & R@50/100    & mR@50/100   & A@50/100    \\ \hline
Motif~\cite{Motif}        & 65.4 / 67.2 & 18.0 / 19.3 & 41.7 / 43.3 & 40.0 / 40.8   & 10.0 / 10.6    & \underline{25.0} / \underline{25.7} & 32.5 / 37.0 & 7.8 / 7.4 & \underline{20.2} / \underline{23.2}  \\
+CogTree~\cite{yu2020cogtree}     & 35.6 / 36.8 & 26.4 / 29.0 & 31.0 / 32.9 & 21.6 / 22.2   & 14.9   / 16.1 & 18.3 / 19.2 & 20.0 / 22.1 & 10.4 / 11.8 & 15.2 / 17.0 \\
+\textcolor{cyan}{DLFE}~\cite{HS1}        & 52.4 / 54.3 & 26.7 / 28.7 & 39.6 / 41.5 & 32.3 / 33.1   & 15.2 / 15.9   & 23.8 / 24.5 & 25.4 / 29.4 & 11.7 / 13.8 & 18.6 / 21.6 \\
+GCL~\cite{T2}         & 42.7 / 44.4 & 36.1 / 38.2 & 39.4 / 41.3 & 26.1 / 27.1   & 20.8 / 21.8   & 23.5 / 24.5 & 18.4 / 22.0 & 16.8 / 19.3 & 17.6 / 20.7 \\
+HML~\cite{HML2022}         & 47.1 / 49.1 & 36.3 / 38.7 & 41.7 / 43.9 & 26.1 / 27.4   & 20.8 / 22.1   & 23.5 / 24.8 & 17.6 / 21.1 & 14.6 / 17.3 & 16.1 / 19.2 \\
+IETrans~\cite{IE} & 54.7 / 56.7 & 30.9 / 33.6 & \underline{42.8} / 45.2 & 32.5 / 33.4 & 16.8 / 17.9& 24.7 / \underline{25.7}& 26.4 / 30.6 &  12.4 / 14.9 & 19.4 / 22.8 \\

+\textcolor{cyan}{RTPB}~\cite{T3}     & 40.3 / 42.6 & 35.4 / 37.4 & 37.9 / 40.0 & 26.0 / 26.9   & 20.0 / 21.0   & 23.0 / 24.0 & 19.0 / 22.5 & 13.1 / 15.5 & 16.1 / 19.0 \\

+PPDL~\cite{ppdl} & 47.2 / 47.6 & 32.2 / 33.3 & 39.7 / 40.5 & 28.4 / 29.3 & 17.5 / 18.2 & 23.0 / 23.8 & 21.2 / 23.9 & 11.4 / 13.5 & 16.3 / 18.7 \\ 
+FGPL~\cite{lyu2022fine} & 51.4 / 55.3  &  33.6 / 38.6  & 42.5 /  \textbf{47.0} & 23.5 / 24.1  &  21.2 / 22.3  & 22.4 / 23.2  & 20.9 / 23.5  & 15.2 / 18.2  &  18.1 / 20.9  \\
+CFA~\cite{CFAl} & 42.3 / 45.1 & 39.9 / 43.0 & 41.1 / 44.1 & 25.7 / 27.4 & 20.9 / 22.4 & 23.3 / 24.9 & 20.7 / 24.4 & 15.3 / 18.1 & 18.0 / 21.3  \\
+Inf~\cite{Pbiswas2023}           & 51.5 / 55.1 & 24.7 / 30.7 & 38.1 / 42.9 & 32.1 / 33.8   & 14.5 / 17.4   & 23.3 / 25.6 & 27.7 / 30.1 & 10.4 / 11.9 & 19.1 / 21.0 \\

\textbf{Our SBG}       & 55.4 / 57.3 & 32.1 / 34.4 & \textbf{43.8} / \underline{45.9} & 34.9 / 35.7   & 17.5 / 18.6   & \textbf{26.2} / \textbf{27.1} & 27.0 / 31.3 & 13.8 / 16.1 &  \textbf{20.4} / \textbf{23.7} \\ \hline

VCtree~\cite{Vctree}       & 65.4 / 67.1 & 17.9 / 19.5 & 41.7 / 43.3 & 46.6 / 47.6 & 12.3 / 13.1 & \underline{29.5} / 30.4 & 31.3 / 35.6 & 7.2 / 8.6 & \underline{19.3} / \underline{22.1}           \\
+CogTree~\cite{yu2020cogtree}       & 44.0 / 45.4 & 27.6 / 29.7 & 35.8 / 37.6 & 30.9 / 31.7   & 18.8 / 19.9   & 24.9 / 25.8 & 18.2 / 20.4 & 10.4 / 12.1 & 14.3 / 16.3 \\
+\textcolor{cyan}{DLFE}~\cite{HS1}          & 51.8 / 53.5 & 25.3 / 27.1 & 38.6 / 40.3 & 33.5 / 34.6   & 18.9 / 20.0   & 26.2 / 27.3 & 22.7 / 26.3 & 11.8 / 13.8 & 17.3 / 20.1 \\
+GCL~\cite{T2}             & 40.7 / 42.7 & 37.1 / 39.1 & 38.9 / 40.9 & 27.7 / 28.7   & 22.5 / 23.5   & 25.1 / 26.1 & 17.4 / 20.7 & 15.2 / 17.5 & 16.3 / 19.1 \\
+HML~\cite{HML2022}         & 47.0 / 48.8 & 36.9 / 39.2 & 42.0 / 44.0 & 27.0 / 28.4   & 25.0 / 26.8   & 26.0 / 27.6 & 17.6 / 21.0 & 13.7 / 16.3 & 15.7 / 18.7 \\
+IETrans~\cite{IE} & 53.0 / 55.0 & 30.3 / 33.9 & 41.7 / 44.5 & 32.9 / 33.8 & 16.5 / 18.1 & 24.7 / 26.0 & 25.4 / 29.3 & 11.5 / 14.0 & 18.5 / 21.7 \\

+\textcolor{cyan}{RTPB}~\cite{T3}    & 41.2 / 43.3 & 33.4 / 35.6 & 37.3 / 39.5 & 28.7 / 30.0   & 24.5 / 25.8   & 26.6 / 27.9 & 18.1 / 21.3 & 12.8 / 15.1 & 15.5 / 18.2 \\
+PPDL~\cite{ppdl} & 47.6 / 48.0 & 33.3 / 33.8 & 40.5 / 40.9 & 32.1 / 33.0 & 21.8 / 22.4 & 27.0 / 27.7 & 20.1 / 22.9 & 11.3 / 13.3 & 15.7 / 18.1 \\
+FPGL~\cite{lyu2022fine} & 42.3 / 43.8 &  37.4 / 40.3  & 39.9 / 42.1  & 27.3 / 28.0 & 26.2 / 27.7  & 26.8 / 27.9 & 20.8 / 23.4  & 15.5 / 18.4 & 18.2 / 20.9 \\
+CFA~\cite{CFAl} & 41.9 / 45.0 & 39.2 / 42.5 & 40.6 / 43.8 & 32.3 / 33.8 & 26.3 / 28.3 & 29.3 / \underline{31.1} & 20.5 / 24.2 & 15.1 / 17.9 & 17.8 / 21.1  \\  
+Inf~\cite{Pbiswas2023} & 59.5 / 60.1 & 28.1 / 30.7 & \underline{43.8} / \underline{45.4} & 40.7 / 41.6 & 17.3 / 19.4 & 29.0 / 30.5 & 27.7 / 30.1& 10.4 / 11.9 & 19.1 / 21.0            \\
\textbf{Our SBG}             & 55.5 / 57.3 & 32.4 / 34.5 & \textbf{44.0} / \textbf{45.9} & 40.8 / 41.9 & 21.8 / 23.1 & \textbf{31.3} / \textbf{32.5} & 26.5 / 30.5  & 12.2 / 14.3 & \textbf{19.4} / \textbf{22.4}           \\ \hline

Transformer~\cite{T1}  & 65.5 / 67.3 & 18.2 / 19.7 & 41.9 / 43.5 & 39.8 / 40.6 & 10.8 / 11.5  & \underline{25.3} / 26.1 & 32.0 / 36.4 & 8.2 / 9.8 & \textbf{20.1} / \underline{23.1}           \\
+CogTree~\cite{yu2020cogtree}      & 38.4 / 39.7 & 28.4 / 31.0 & 33.4 / 35.4 & 22.9 / 23.4   & 15.7 / 16.7   & 19.3 / 20.1 & 19.5 / 21.7 & 11.1 / 12.7 & 15.3 / 17.2 \\

+HML~\cite{HML2022}          & 45.6 / 47.8 & 33.3 / 35.9 & 39.5 / 41.9 & 22.5 / 23.8   & 19.1 / 20.4   & 20.8 / 22.1 & 15.0 / 17.7 & 15.0 / 17.7 & 15.0 / 17.7 \\
+PPDL~\cite{ppdl} & 46.5 / 47.2 & 35.7 / 36.0 & 36.4 / 41.6& 28.5 / 29.3 & 17.9 / 18.8 & 23.2 / 24.1 & 20.7 / 24.2 & 11.5 / 13.2 & 16.1 / 18.6        \\
+FPGL~\cite{lyu2022fine}& 50.9 / 54.6 &  36.3 / 40.1 & \underline{43.6} / \textbf{47.4} & 21.3 / 22.1 & 22.5 / 24.1 & 21.9 / 23.1 & 19.1 / 21.8 & 17.5 / 20.5 & 18.3 / 21.2  \\
+CFA~\cite{CFAl}  & 46.2 / 48.9 & 38.6 / 41.5 & 42.4 / 45.2 & 29.6 / 28.1 & 20.9 / 22.7  & 24.5 / 26.2 & 21.0 / 24.7 & 15.0 / 17.9 & 18.0 / 21.3         \\
+IETrans~\cite{IE} & 51.8 / 53.8 & 30.8 / 34.7 & 41.3 / 44.3 & 32.6 / 33.5 & 17.4 / 19.1 & 25.0 / \underline{26.3} &  25.5 / 29.6 & 12.5 / 15.0 & \underline{19.0} / 22.3 \\

\textbf{Our SBG}     &     55.8 / 57.6 & 33.3 / 35.7 & \textbf{44.6} / \underline{46.7} & 35.6 / 36.5   & 18.5 / 19.4   & \textbf{27.1} / \textbf{28.0} & 24.5 / 28.5  & 15.6 / 18.0 & \textbf{20.1} / \textbf{23.3} \\ 

\hline

\end{tabular}
}

	\label{tab:alltotal}

\end{table*}

\subsection{Experiments on VG}
\noindent\textbf{Comparison with the State-of-the-Art Two-Stage Methods on VG.} We compare our SBG with the state-of-the-art two-stage methods on three mainstream SGG models: Motif~\cite{Motif}, VCtree~\cite{Vctree}, and Transformer~\cite{T1}, as illustrated in \cref{tab:alltotal}. For Motif, VCtree, and Transformer, our SBG improves significantly all mR@50/100, demonstrating our SBG's strong detection capability for tail relationships. Compared to the state-of-the-art two-stage methods, our SBG outperforms all other methods in terms of A@50/100, exhibiting a better-balanced improvement between R@50/100 and mR@50/100. Specifically, considering the PredCls of Transformer as an example, other methods sacrifice R@50/100 significantly while improving mR@50/100. By contrast, our SBG minimizes sacrifices on R@50/100 while maintaining a competitive performance on mR@50/100. These results not only demonstrate the effectiveness of our SBG but also show our SBG's capacity to improve tail relationships while avoiding excessive suppression of head relationships. Furthermore, our SBG is mostly highest in terms of A@K on three mainstream models, but the performance of the other methods is inconsistent (i.e., no method has equal competitiveness on three models), which reflects the strong generalization of our SBG. 

\begin{table*}[!ht]
	\centering
    \small
     \setlength{\tabcolsep}{1mm}
   	\caption{Improvement comparison of dataset-level bias correction methods DLFE~\cite{HS1}, RTPB~\cite{T3} on PredCls, SGCls, and SGDet tasks of Motif and VCtree models for R@100 (\%), mR@100 (\%), and A@100 (\%). Imp (\%) indicates the degree of improvement for Motif and VCtree models. The underlined values have the same meaning as in \cref{tab:alltotal}.
	}
    \resizebox{\linewidth}{!}{
	\begin{tabular}{c|ccc|ccc|ccc}
\hline
 & \multicolumn{3}{c|}{PredCls}             & \multicolumn{3}{c|}{SGCls}                   & \multicolumn{3}{c}{SGDet}               \\ \hline
Model          & R@100/Imp   & mR@100/Imp   & A@100/Imp    & R@100/Imp      & mR@100/Imp     & A@100/Imp    & R@100/Imp    & mR@100/Imp   & A@100/Imp    \\ \hline
Motif~\cite{Motif}        &  67.2 &  19.3 &  43.3 &  40.8   &  10.6    & 25.7 &  37.0 &  7.4 & 23.2  \\
+DLFE~\cite{HS1}        &  54.3 / $-$\underline{12.9} &  28.7 / +9.4 & 41.5 / $-$\underline{1.8} &  33.1 / $-$\underline{7.7}   & 15.9 / +5.3   &  24.5 / $-$\underline{1.2} & 29.4 / $-$\underline{7.6} &  13.8 / +6.4 &  21.6 / $-$\underline{1.6} \\
+RTPB~\cite{T3}     & 42.6 / $-$24.6  & 37.4 / +\textbf{18.1} &  40.0 / $-$3.3 &  26.9 / $-$13.9   &  21.0 / +\textbf{10.4}   & 24.0 / $-$1.7 &  22.5 / $-$14.5 &  15.5 / +\underline{8.1} &  19.0 / $-$4.2 \\
\textbf{Our SBG}       &  57.3 / $-$\textbf{9.9} &  34.4 / +\underline{15.1} &  45.9 / +\textbf{2.6} &  35.7 / $-$\textbf{5.1}  &  18.6 / +\underline{8.0}   &  27.1 / +\textbf{1.4}&  31.3 / $-$\textbf{5.7} & 16.1 / +\textbf{8.7} &  23.7 / +\textbf{0.4} \\ \hline
VCtree~\cite{Vctree}       &  67.1 &  19.5 & 43.3 &  47.6 &  13.1 & 30.4 &  35.6 &  8.6 &  22.1          \\
+DLFE~\cite{HS1}          & 53.5 / $-$\underline{13.6} &  27.1 / +7.6 & 40.3 / $-$\underline{3.0} &  34.6 / $-$\underline{13.0}   &  20.0 / +6.9   &  27.3 / $-$3.1 &  26.3 / $-$\underline{9.3} &  13.8 / +5.2 &  20.1 / $-$\underline{2.0} \\
+RTPB~\cite{T3}    &  43.3  / $-$17.7 & 35.6 / +\textbf{16.1} & 39.5 / $-$3.8 &  30.0 / $-$17.6  &  25.8 / +\textbf{12.7}   &  27.9  / $-$\underline{2.5}& 21.3 / $-$14.3 &  15.1 / +\textbf{6.5} & 18.2 / $-$3.9 \\
\textbf{Our SBG}             & 57.3 / $-$\textbf{9.8} &  34.5 / +\underline{15.0} &  45.9 / +\textbf{2.6}  &  41.9 / $-$\textbf{5.7} &  23.1 / +\underline{10.0} &  32.5 / +\textbf{2.1} & 30.5 / $-$\textbf{5.1}  & 14.3 / +\underline{5.7} &  22.4 / +\textbf{0.3}          \\ \hline
\end{tabular}
}

	\label{tab:a2}

\end{table*}

\begin{figure}[h]
\centering  
\begin{minipage}{0.45\textwidth} 
    \centering  
    \includegraphics[width=0.9\linewidth]{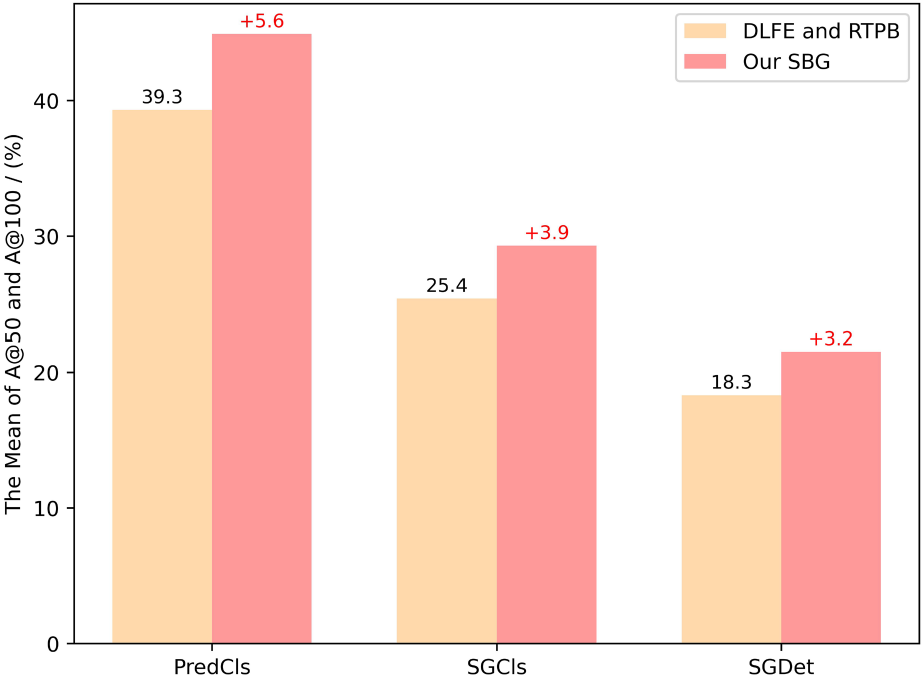}
    \caption{Comparison with DLFE and RTPB on the overall performance.}
    \label{fig:RBP}
\end{minipage}
\hspace{0.05\textwidth}  
\begin{minipage}{0.45\textwidth}  
\centering 
    \small

    \setlength{\tabcolsep}{1mm}
            \captionof{table}{Efficiency comparison with DLFE~\cite{HS1} and RTPB~\cite{T3}. $\uparrow$ and $\downarrow$ denote the percentage change of our SBG compared to DLFE and RTPB.}
    
    \resizebox{\linewidth}{!}{
    
        \begin{tabular}{c|cccc}
        \hline
        Method &A@100   & Training Time  & Inference Speed     & Parameters     \\ \hline
        DLFE~\cite{HS1}      &  41.5 & 22.9   & 0.1049      &  253.41 \\
        \textbf{Our SBG}       & 45.9 \textcolor{darkgreen}{$\uparrow$10.6\%} & 12.6 \textcolor{darkgreen}{$\downarrow$45.0\%} & 0.1062 \textcolor{darkgreen}{$\uparrow$1.2\%}    & 254.75 \textcolor{darkgreen}{$\uparrow$0.5\%} \\ \hline
        RTPB~\cite{T3}     &40.1 & 12.1 & 0.1039     & 254.05  \\
        \textbf{Our SBG}       &45.9 \textcolor{darkgreen}{$\uparrow$14.5\%} &  12.6 \textcolor{darkgreen}{$\uparrow$4.1\%} & 0.1062 \textcolor{darkgreen}{$\uparrow$2.2\%}    & 254.75 \textcolor{darkgreen}{$\uparrow$0.3\%}  \\ \hline
        \end{tabular}
    }

    \label{tab:TTT}
\end{minipage}
\end{figure}

\noindent\textbf{Superiority of Sample-Level Bias Correction.} To demonstrate the superiority of sample-level bias correction, we compare our SBG with dataset-level bias correction methods (light blue in \cref{tab:alltotal}) DLFE and RTPB on VG. Compared to DLFE, our method outperforms on R@50/100, mR@50/100, and A@50/100, indicating its superior performance. Although our mR@50/100 metrics (excluding SGDet task of Motif) are slightly lower than RTPB, we achieve a higher R@50/100 and A@50/100. These demonstrate that our SBG exhibits a better overall performance. To provide a comprehensive analysis of the balance between R@50/100 and mR@50/100, we compare the improvements on Motif and VCtree with RTPB and DLFE, as depicted in \cref{tab:a2}. We have the smallest decrease on all R@100 metrics while realizing a competitive improvement on all mR@100 metrics, indicating a more balanced improvement across the R@50/100 and mR@50/100 metrics. Additionally, we compare our SBG on the overall performance (i.e., the mean of A@50 and A@100) with DLFE and RTPB across PredCls, SGCls, and SGDet tasks on Motif and VCtree models, as illustrated in \cref{fig:RBP}. The results indicate that our SBG achieves the improvements of 5.6\%, 3.9\%, and 3.2\% across the three tasks, further demonstrating the superiority of our SBG (i.e., the sample-level bias correction).

\cref{tab:TTT} presents the efficiency comparison on the PredCls task of Motif model for Training Time (h), Inference Speed (s/img), and Parameters (M). Compared to DLFE, we achieve a 10.6\% improvement in A@100, while reducing training time by 45.0\%. However, the inference speed and parameters only increase by 1.2\% and 0.5\%, respectively. Compared to RTPB, we achieve a 14.5\% improvement in A@100, with only marginal increases of 4.1\% in training time, 2.2\% in inference speed, and 0.3\% in parameters. These results are deemed acceptable, indicating that our SBG achieves a significant A@100 enhancement with minimal efficiency losses.

\noindent\textbf{Extensibility to One-Stage SGG Methods.} We validate our SBG on one-stage SGG methods on SGDet task which they support of VG for R@50/100, mR@50/100, and A@50/100 to show the expansibility in \cref{tab:all}. From the table, we find that our SBG has a significant improvement on mR@50/100, demonstrating our SBG's strong detection capability for tail relationships. While improving mR@50/100, there is a lesser sacrifice on R@50/100, thereby demonstrating the superior comprehensive performance. The results validate the extensibility of our SBG.

\begin{table*}[!htb]
\centering
\begin{minipage}{.48\textwidth}
\centering
\small
\setlength{\tabcolsep}{1mm}
\caption{Performance on one-stage SGG methods.}
\resizebox{0.8\linewidth}{!}{
\begin{tabular}{c|ccc}
\hline
 & \multicolumn{3}{c}{SGDet}               \\ \hline
Model & R@50/100    & mR@50/100   & A@50/100    \\ \hline
ISG~\cite{ISG}  &30.8 / 35.6 & 19.5 / 23.4 & 25.2 / 29.5\\ 
\textbf{Our SBG} & 29.3 / 33.6  & 24.1 / 28.2& \textbf{26.7} / \textbf{30.9} \\ \hline
SGTR~\cite{li2022sgtr}  & 24.6 / 28.4 & 12.0 / 15.2 & 18.3 / 21.8 \\ 
+FGPL-A~\cite{lyu2023adaptive} & 24.6 / 28.2 & 12.5 / 19.4 & 18.6 / \textbf{23.8} \\
\textbf{Our SBG}  & 22.3 / 26.3 &  18.1 / 20.9 & \textbf{20.2} / 23.6\\ \hline
SS R-CNN~\cite{SSR-CNN}  & 33.5 / 38.4 & 8.6 / 10.3 & 21.1 / 24.4 \\
+FGPL-A~\cite{lyu2023adaptive} & 31.9 / 36.6 & 9.5 / 11.9 & 20.7 / 24.3 \\ 
\textbf{Our SBG}  & 27.3 / 31.1 & 20.0 / 23.5 & \textbf{23.7} / \textbf{27.3} \\ \hline
\end{tabular}
}
\label{tab:all}
\end{minipage}%
\hfill
\begin{minipage}{.48\textwidth}
\centering
\small
\setlength{\tabcolsep}{1mm}
\caption{Comparison on Motif for the PredCls task.}
\label{tab:three}
\resizebox{\linewidth}{!}{

\begin{tabular}{c|cc|cc|ccc}
\hline

 & \multicolumn{2}{|c|}{GQA}             & \multicolumn{2}{c|}{VG}                   & \multicolumn{3}{c}{VG-1800}               \\ \hline
Model       & A@50   & A@100     & A@50   & A@100   & F-Acc(Top-1) & F-Acc(Top-5) & F-Acc(Top-10)      \\ \hline
Motif  & 45.4 & 46.7  & 41.7 & 43.3  & 1.21 & 5.20  & 8.33    \\
+CFA&  47.2 & 48.6 & 41.1 & 44.1  & 5.33 & 18.25   & 24.51  \\
Our SBG  &  47.4\textcolor{darkgreen}{$\uparrow$0.2\%} &  48.7\textcolor{darkgreen}{$\uparrow$0.1\%} & 43.8\textcolor{darkgreen}{$\uparrow$2.7\%} & 45.9\textcolor{darkgreen}{$\uparrow$1.8\%} & 13.45\textcolor{darkgreen}{$\uparrow$8.12\%}     & 26.74\textcolor{darkgreen}{$\uparrow$8.49\%} & 34.99\textcolor{darkgreen}{$\uparrow$10.48\%}  \\ \hline

\hline
\end{tabular}
}
\label{table:local_gloabal}
\end{minipage}
\end{table*}

\begin{table*}[!]

	\caption{Performance comparison with RTPB~\cite{T3}, CFA~\cite{CFAl} on PredCls, SGCls, and SGDet tasks of GQA for R@50/100 (\%), mR@50/100 (\%), and A@50/100 (\%). The underlined values has the same meaning as in \cref{tab:alltotal}.
	}
	\centering

    \small
     \setlength{\tabcolsep}{1mm}

    \resizebox{\linewidth}{!}{
	\begin{tabular}{c|ccc|ccc|ccc}
\hline                   
            & \multicolumn{3}{c|}{PredCls}             & \multicolumn{3}{c|}{SGCls}                   & \multicolumn{3}{c}{SGDet}               \\ \hline
 Model          & R@50/100    & mR@50/100   & A@50/100    & R@50/100      & mR@50/100     & A@50/100    & R@50/100    & mR@50/100   & A@50/100    \\ \hline
Motif~\cite{Motif}  & 66.2 / 67.6 &  24.5 / 25.8 & 45.4 / 46.7 &  33.4 / 34.0 & 12.0 / 12.5 & 22.7 / 23.3 & 30.3 / 34.4 & 10.5 / 12.2 &  20.4 / 23.3 \\
+RTPB~\cite{T3}   & 50.1 / 51.5  & 43.9 / 45.2 & 47.0 / 48.4 & 25.7 / 26.3 & 21.3 / 21.7& \underline{23.5} / 24.0  & 21.9 / 25.6& 21.0 / 23.2 & \textbf{21.5} / \underline{24.4} \\
+CFA~\cite{CFAl} & 50.1 / 51.2 & 44.3 / 46.0 &  \underline{47.2} / \underline{48.6} & 23.6 / 24 & 23.4 / 24.2  & \underline{23.5} / \underline{24.1} & 19.8 / 23.9 & 22.8 / 24.7 & \underline{21.3} / 24.3 \\
\textbf{Our SBG} & 56.6 / 58.0 & 38.1 / 39.3 &  \textbf{47.4} / \textbf{48.7} & 28.6 / 29.2 & 18.8 / 19.4  & \textbf{23.7} / \textbf{24.3}   &25.2 / 29.4 & 17.8 / 19.6 & \textbf{21.5} / \textbf{24.5}  \\ \hline

VCtree~\cite{Vctree}   & 65.5 / 67.1 & 24.8 / 26.3  & 45.5 / 46.7 & 33.5 / 34.1& 12.5 / 13.0 & 23.0 / 23.6 & 28.4 / 32.1 & 9.9 / 11.7 & 19.2 / 21.9 \\
+RTPB~\cite{T3}   & 49.9 / 51.4  & 42.6 / 44.0  & 46.3 / 47.7 & 26.0 / 26.6  & 22.2 / 22.6 & 24.1 / 24.6 &20.6 / 24.0  & 20.0 / 22.3  & 20.3 / \underline{23.2}  \\
+CFA~\cite{CFAl} & 48.7 / 49.3 & 45.1 / 47.5 & \underline{46.9} / \underline{48.4} & 26.1 / 26.7  & 22.9 / 23.3  & \underline{24.5} / \underline{25.0}& 18.9 / 23.0 & 21.9 / 23.4 & \underline{20.4} / \underline{23.2} \\ 
\textbf{Our SBG}& 57.9 / 59.4 & 37.2 / 38.5 & \textbf{47.6} / \textbf{49.0} & 29.7 / 30.3  & 19.5 / 20.0  & \textbf{24.6} / \textbf{25.2}   &  24.3 / 27.8  & 16.6 / 18.8 &  \textbf{20.5} / \textbf{23.3}  \\ \hline
Transformer~\cite{T1}  & 67.5 / 68.9 & 26.8 / 28.2 & 47.2 / 48.6 & 34.6 / 35.1 & 14.7 / 15.2   & \underline{24.7} / \underline{25.2} & 30.6 / 34.7 & 12.2 / 14.1 & 21.4 / 24.4  \\
+RTPB~\cite{T3} &50.8 / 52.3  & 44.6 / 45.8  & 47.7 / 49.1 & 26.0 / 26.6 & 21.9 / 22.3  & 24.0 / 24.5 & 21.1 / 24.8& 21.3 / 23.6&  21.2 / 24.2  \\
+CFA~\cite{CFAl}  & 50.5 / 52.8 & 46.1 / 47.2 & \underline{48.3} / \underline{50.0}  & 24.9  / 25.5  & 23.5 / 24.1 & 24.2 / 24.8 & 22.0 / 26.0 & 21.4 / 23.6 &  \underline{21.7} / \underline{24.8}    \\
\textbf{Our SBG}        & 58.6 / 60.0  & 41.6 / 42.9    & \textbf{50.1} / \textbf{51.5}   & 28.6 / 29.2    & 21.0 / 21.6    & \textbf{24.8} / \textbf{25.4}  & 24.3 / 28.0  & 20.3 / 22.6   & \textbf{22.3} / \textbf{25.3}   \\ \hline
\end{tabular}
}

\label{table: GQA}
\end{table*}

\noindent\textbf{Comparison with the State-of-the-Art Two-Stage Methods on GQA and VG-1800.} 
Following VG, we validate the effectiveness of our SBG on GQA as shown in \cref{table: GQA}. We compare the best dataset-level bias correction method RTPB~\cite{T1} and the latest method CFA~\cite{CFAl}. From the results, it can be seen that our SBG achieves a significant improvement for all baselines on mR@50/100 and outperforms RTPB and CFA on almost all A@50/100. Furthermore, we compare the gains of our SBG with CFA in \cref{tab:three} on three datasets GQA, VG, and VG-1800, where their data volume increase and the long-tailed effect is more severe sequentially. The results show that the gains of our SBG are more significant on larger datasets with more severe long-tailed effects. The trivial gains on GQA in \cref{table: GQA} are due to that compared to VG and VG-1800, the data volume of GQA is smaller and the long-tailed effect of GQA is weaker.

\noindent\textbf{Ablation Studies.}  

\textbf{i) The Effect of Region Scope for Sample-Level Bias Prediction:} In our SBG, the sample-specific biases are predicted by utilizing the contextual information from object pairs' union regions which is abbreviated as $union$. We expand the region scope of contextual information to the entire image which is abbreviated as $entire$ for comparison on PredCls task using of Transformer model, as shown in \cref{table: fff }. The A@50/100 decreases when utilizing the contextual information of the entire image, indicating that the contextual information of the entire image introduces additional interference. This is because, in reality, some object pairs' union regions have already a sufficiently large scope that can provide enough contextual information.

\begin{table}[!htb]

\begin{minipage}{.45\linewidth}
\caption{The effect of region scope for sample-level bias prediction.}
\centering
\small
\setlength{\tabcolsep}{1mm}
\resizebox{\linewidth}{!}{

\begin{tabular}{c|ccc}
\hline
        &\multicolumn{3}{c}{PredCls}    \\  \hline
  & {R@50/100}    & mR@50/100   & A@50/100    \\  \hline
$entire$ & 48.0 / 48.8 & 32.6 / 35.1 & 40.3 / 42.0 \\
$union$ & 55.8 / 57.6 & 33.3 / 35.7 & \textbf{44.6} / \textbf{46.7} \\   \hline
\end{tabular}%
}
\label{table: fff }
\end{minipage}\hfill
\begin{minipage}{.45\linewidth}
\caption{The effect of global bias $\mathbf{b}^{glo}$, $\checkmark$ indicates the use of $\mathbf{b}^{glo}$ or $f_{uni}$.}
\centering
\small
\setlength{\tabcolsep}{1mm}
\resizebox{\linewidth}{!}{%
\begin{tabular}{cc|ccc}
\hline
 \multicolumn{2}{c|}{} & \multicolumn{3}{c}{PredCls}             \\ \hline
 $f_{uni}$ & $\mathbf{b}^{glo}$ & R@50/100    & mR@50/100   & A@50/100    \\ \hline
   & & 65.5 / 67.3 & 18.2 / 19.7 & 41.9 / 43.5     \\
 & $\checkmark$ & 65.3 / 67.0  & 18.9 / 20.1   & 42.1 / 43.5  \\
$\checkmark$ &    &54.7 / 55.9 & 30.4 / 33.7 & 42.6 / 44.8  \\
$\checkmark$ &$\checkmark$ &55.8 / 57.6 & 33.3 / 35.7 & \textbf{44.6} / \textbf{46.7}   \\ \hline
\end{tabular}%
}
\label{tab:ffff}
\end{minipage} 
\end{table}

\textbf{ii) The Effect of Global Bias $\mathbf{b}^{glo}$:} In \cref{sec:WGAN}, we utilize the global bias $\mathbf{b}^{glo}$ in the prediction of sample-specific biases, and the impact of it on the Predcls task of Transformer model is analyzed in \cref{tab:ffff}, $f_{uni}$ denotes object pairs’ union features.
The introduction of global bias $\mathbf{b}^{glo}$ results in an improvement, as indicated by rows 3 and 4 in the table, showing its effect on improving the accuracy of bias correction. However, we can observe that using $\mathbf{b}^{glo}$ only is not satisfactory.  As it is a set of fixed values, the training has minimal effectiveness. Additionally, the union features $f_{uni}$ have a crucial impact, as our SBG relies heavily on the object pairs' union features to predict the sample-specific biases. More ablation experiments are in Supplementary Materials.

\begin{figure}[h]
    \begin{center}
        \includegraphics[width=0.75\columnwidth]{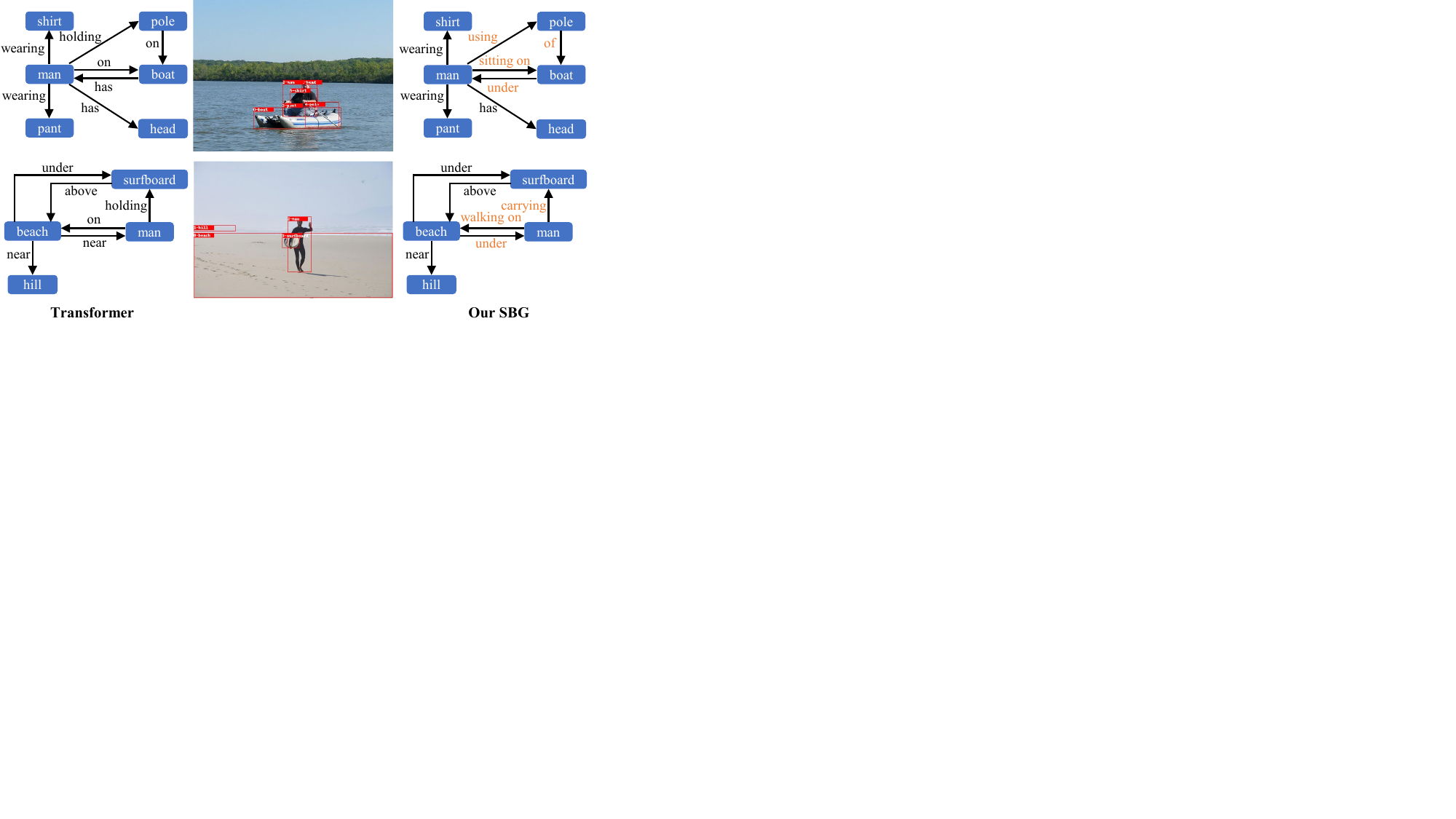}
    \end{center}
    
    \caption{Comparison of the generated scene graphs between Transformer and SBG.}
    \label{fig:vis_fine-graine}
\end{figure}

\noindent\textbf{Visualization Results.}
\cref{fig:vis_fine-graine} presents the results of SGG from Transformer model and our SBG. The left side displays the scene graph generated by Transformer model, while the right side displays the scene graph generated by SBG. The differences in relationships between the two scene graphs are highlighted in the orange font. It is evident that the relationships obtained by Transformer model consist mostly of the coarse-grained relationships, such as ``\emph{on}'', ``\emph{has}'', and ``\emph{near}''. By contrast, our SBG predicts more fine-grained relationships, such as ``\emph{sitting on}'', ``\emph{walking on}'', and ``\emph{carrying}''. 

\noindent\textbf{Generalization for Object Detection.} We extend our SBP to other tasks with the long-tailed problem, such as object detection in Tab.~\ref{tab:obj}. We experiment on the COCO dataset~\cite{lin2014microsoft}, which contains 80 object categories with the severe long-tailed effect. After applying our SBP, the $mAP$ achieves 1.2\% improvements, with the tail categories achieving a larger 2.8\% improvements. 

\begin{table*}[!ht]
     \caption{Generalization for object detection. $mAP_{tail}$ is $mAP$ for the last 50 categories of COCO~\cite{lin2014microsoft} ordered by long tail.}
     
	\label{tab:obj}
\centering
    \small
     \setlength{\tabcolsep}{1mm}
    \resizebox{0.6\linewidth}{!}{
	\begin{tabular}{c|ccc}
\hline
Model (backbone, training schedule) & $mAP$ & $mAP_{tail}$ & \\ \hline
Faster R-CNN~\cite{ren2015faster} (R-50-FPN, 1x)  & 36.4 & 41.6 &  \\
Faster R-CNN~\cite{ren2015faster} (R-50-FPN, 1x) + SBP & 37.6 \textcolor{darkgreen}{$\uparrow$1.2\%} & 44.4 \textcolor{darkgreen}{$\uparrow$2.8\%} &  \\
\hline
\end{tabular}
}
    
\end{table*}

\section{Conclusion \& Future Work}
In this paper, to tackle the long-tailed problem, we propose one novel method SBP to conduct sample-level bias correction, and further generate the fine-grained scene graph. Specifically, we design a BGAN to predict the sample-specific bias. Extensive experiments on VG, GQA, and VG-1800 datasets validate the effectiveness and generalizability of our SBG. We believe that this work contributes to the advancement of research in this field and offers insights into tackling the long-tailed issue. Our future work aims to further enhance the performance of the proposed method and extend its applicability to other tasks.

\section*{Acknowledgements}
This work was partly supported by the National Natural Science Foundation of China (42371321 and 42030102), Natural Science Foundation of HuBei Province (Grant No. 2024AFB283), and Science Foundation of China Three Gorges University (Grant No. 2023RCKJ0022).

\bibliographystyle{splncs04}
\bibliography{main}

\appendix
\section{Dataset Details}

GQA~\cite{hudson2019gqa} is a vision-and-language dataset, consisting of a total of 113k images. We retain images only with the most frequent 160 object and 60 relationship categories for experiments. Then it contains 59,588 images, of which 41,773 (70\%) images are used for the training, and 17,815 (30\%) images are used for the testing. We follow ~\cite{T1} to sample a 5k validation set from the training set for parameter tuning. The detailed list of the most frequent 160 object and 60 relationship categories is shown in \cref{table:type_vg}.

\begin{table}[hpb]
\caption{List of object and relationship categories in GQA.}
\resizebox{\columnwidth}{!}{
\begin{tabular}{c|l}
\hline
& \multicolumn{1}{c}{\textbf{categories}}                                                                                 \\ \hline
\textbf{object}     & \multicolumn{1}{l}{\begin{tabular}[c]{@{}l@{}}
'window', 'man', 'shirt', 'tree', 'wall', 'person', 'building', 'ground', 'sky', 'sign', 'head',\\
'pole', 'hand', 'grass', 'hair', 'leg', 'car', 'woman', 'leaves', 'trees', 'table', 'ear', 'pants',\\
'people', 'eye', 'water', 'door', 'fence', 'nose', 'wheel', 'chair', 'floor', 'arm', 'jacket', \\
'hat', 'shoe', 'tail', 'clouds', 'leaf', 'face',' letter', 'plate', 'number', 'windows', 'shorts',\\
'road', 'flower', 'sidewalk', 'bag', 'helmet', 'snow', 'rock', 'boy', 'cloud', 'tire', 'logo', \\
'roof', 'glass', 'street', 'foot', 'umbrella', 'legs', 'post', 'jeans', 'mouth', 'boat', 'cap',\\
'bottle', 'bush', 'girl', 'flowers', 'shoes', 'picture', 'glasses', 'field', 'mirror', 'bench',\\
'box', 'dirt', 'bird', 'clock', 'neck', 'bowl', 'food', 'bus', 'letters', 'pillow', 'shelf',\\
'train', 'trunk', 'horse', 'airplane', 'plant', 'coat', 'lamp', 'kite', 'wing', 'elephant', 'house',\\
'cup', 'paper', 'dog', 'seat', 'sheep', 'street light', 'counter', 'branch', 'glove', 'banana',\\
'giraffe', 'book', 'rocks', 'cow', 'truck', 'racket', 'ceiling', 'flag', 'skateboard', 'cabinet',\\
'zebra', 'eyes', 'ball', 'bike', 'wheels', 'sand', 'surfboard', 'frame', 'hands', 'motorcycle',\\
'feet', 'windshield', 'finger', 'bushes', 'player', 'child', 'hill', 'sink', 'bed', 'cat', 'container',\\
'traffic light', 'sock', 'tie', 'towel', 'pizza', 'paw', 'backpack', 'collar', 'basket', 'mountain',\\
'vase', 'lid', 'phone', 'branches', 'animal', 'donut', 'fur', 'license plate', 'laptop', 'lady'\\
\end{tabular}} \\ 
\hline
\textbf{relationship} & \begin{tabular}[c]{@{}l@{}}
'on', 'wearing', 'of', 'near', 'in', 'behind', 'in front of', 'holding', 'on top of', 'next to', \\
'above', 'with', 'below', 'by', 'sitting on', 'under', 'on the side of', 'beside', 'standing on',\\
'inside', 'carrying', 'at', 'walking on', 'riding', 'standing in', 'around', 'covered by', 'hanging on',\\
'lying on', 'eating', 'watching', 'looking at', 'covering', 'sitting in', 'on the front of', \\
'hanging from', 'parked on', 'riding on', 'using', 'covered in', 'flying in', 'sitting at', 'walking in',\\
'playing with', 'full of', 'filled with', 'on the back of', 'crossing', 'swinging', 'surrounded by',\\
'standing next to', 'reflected in', 'covered with', 'contain', 'touching', 'pulling', 'pulled by', \\
'flying', 'leaning on', 'hitting'
\end{tabular} \\ \hline
   
\end{tabular}
}

\label{table:type_vg}
\end{table}

\begin{figure*}[!htp]
    \begin{center}
        \includegraphics[width=0.9\columnwidth]{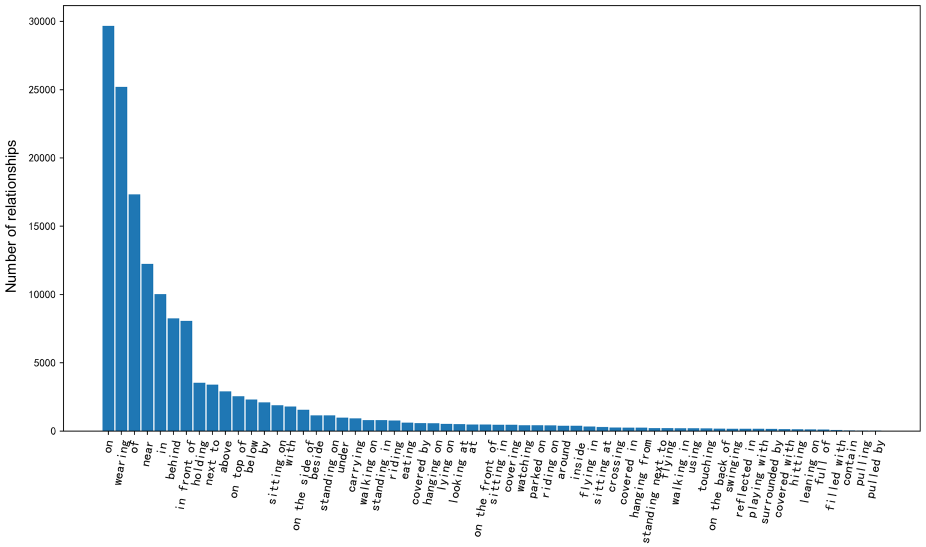}
    \end{center}
    
    \caption{Quantity distribution for each relationship varies from many to few.}
    \label{fig:vis}
    
\end{figure*}

We visualize the quantity distribution for each relationship as shown in \cref{fig:vis}, GQA exhibits a severe long-tailed effect, with a highly imbalanced distribution between head categories (e.g., ``\emph{on}'', ``\emph{wearing}'', ``\emph{of}'') and tail categories (e.g., ``\emph{contain}'', ``\emph{pulling}'', ``\emph{pulled by}'').

\section{Ablation Studies} 
\textbf{iii) The Effect of Weight Factor $\alpha$:} To assess the impact of $\alpha$ for SBG, we conduct the PredCls task on Transformer model. We validate a range of values (0.050, 0.075, 0.100) for $\alpha$. The performance is presented in \cref{table:a factor}. From the results, it can be observed that the A@50/100 metric achieves the highest performance when $\alpha$ is set to 0.075, indicating the optimal performance of SBG.
\begin{table}[!htb]
\caption{The effect of weight factor $\alpha$.}
 \small
\centering
\setlength{\tabcolsep}{1mm}
\begin{tabular}{c|ccc}
\hline
        &\multicolumn{3}{c}{PredCls}    \\  \hline
  Weight Factor $\alpha$ & {R@50/100}    & mR@50/100   & A@50/100    \\  \hline
0.050 & 55.9 / 57.7 & 33.0 / 35.3 & 44.5 / 46.5    \\
0.075 & 55.8 / 57.6 & 33.3 / 35.7 & \textbf{44.6} / \textbf{46.7} \\ 
0.100 & 57.2 / 59.0  & 32.0 / 34.1  & \textbf{44.6} / 46.6 \\
\hline

\end{tabular}

\label{table:a factor}
\end{table}

\textbf{iV) The Effect of Training Mode:} In Section \textcolor{red}{3.2}, we employ a gradual training mode, where the parameters of the classic SGG model are frozen after the training, and subsequently, the training of BGAN is conducted. The comparison between the gradual training and integrated training of SBG on the PredCls task of Transformer model is presented in \cref{table:C_G_IT}. The results indicate that the gradual training outperforms the integrated training. This is because SBG is trained based on the output of the classic SGG model. However,
the output of the classic SGG model using the integrated training is continuously varied, thus leading to the unstable training for SBG.

\begin{table}[!htb]
\caption{The effect of training mode.}
 \small
\centering
\setlength{\tabcolsep}{1mm}
\begin{tabular}{c|ccc}
\hline
        &\multicolumn{3}{c}{PredCls}    \\  \hline
  Training Mode & {R@50/100}    & mR@50/100   & A@50/100    \\  \hline
integrally & 56.4 / 58.1 & 32.6 / 34.8 & 44.5 / 46.5   \\
gradually  & 55.8 / 57.6 & 33.3 / 35.7 & \textbf{44.6} / \textbf{46.7} \\  \hline

\end{tabular}

\label{table:C_G_IT}
\end{table}

\textbf{V) The Superiority of BGAN for Sample-Level Bias Prediction:} To demonstrate 
the sample-level bias's prediction capability of BGAN, which employs the one-dimensional convolution network, we conduct a comparison involving three networks: a conventional 5-layer fully connected network (denoted as $FC_{5}$), a 5-layer one-dimensional convolutional network (denoted as $1D_{5}$), and a fully connected BGAN (denoted as ${\rm BGAN}_{FC}$), on the Predcls task of Transformer model. The results are presented in \cref{table:WGANand5}. It can be observed that in the case of $FC_{5}$ and $1D_{5}$ networks, the $1D_{5}$ network outperforms the $FC_{5}$ network slightly, as the $1D_{5}$ network benefits from the translation invariance and strong local receptive field provided by one-dimensional convolutions. Similarly, the performance of the BGAN based on one-dimensional convolutions is slightly better than that of ${\rm BGAN}_{FC}$ which uses the fully connected networks. Furthermore, by comparing the first and last two rows, BGAN exhibits stronger capabilities for the sample-level bias prediction than conventional neural networks.

\begin{table}[!htb]
\caption{The superiority of BGAN for sample-level bias prediction.}
 \small
\centering
\setlength{\tabcolsep}{1mm}
\begin{tabular}{c|ccc}
\hline
        &\multicolumn{3}{c}{PredCls}    \\  \hline
  Network & {R@50/100}    & mR@50/100   & M@50/100    \\  \hline
$FC_{5}$& 42.3 / 44.0 & 37.4 / 39.7 & 39.9 / 41.9   \\
$1D_{5}$& 41.9 / 43.7 & 38.1 / 40.2 & 40.0 / 42.0   \\
${\rm BGAN}_{FC}$ & 60.1 / 62.9 & 28.4 / 30.0 & 44.3 / 46.5 \\
$\rm BGAN$ & 55.8 / 57.6 & 33.3 / 35.7 & \textbf{44.6} / \textbf{46.7}   \\ \hline

\end{tabular}

\label{table:WGANand5}
\end{table}

\textbf{Vi) The Analysis for Feature Mapping $\phi$:} In Section \textcolor{red}{3.2}, when constructing the correction bias set, we utilize an encoder that includes a single layer of transformer (denoted as $Trans_{1}$) to map high-dimensional features to one-dimensional features. We compare this approach with a conventional fully connected mapping (denoted as $FC$) and an encoder containing two layers of transformer (denoted as $Trans_{2}$), based on the Predcls task of Transformer model. The results are presented in Table \cref{table:mapping}. It is evident that using $Trans_{1}$ for feature mapping yields the best performance. Compared to $FC$, $Trans_{1}$ demonstrates superior performance by leveraging the strong interaction capabilities of the transformer. Moreover, $Trans_{2}$ is relatively complex and results in a performance decline.

\begin{table}[!htb]
\caption{The analysis for feature mapping $\phi$.}
 \small
\centering
\setlength{\tabcolsep}{1mm}
\begin{tabular}{c|ccc}
\hline
        &\multicolumn{3}{c}{PredCls}    \\  \hline
  Mapping Method & {R@50/100}    & mR@50/100   & M@50/100    \\  \hline
$FC$& 55.8 / 57.8 & 32.1 / 34.7 & 44.0 / 46.3   \\
$Trans_{1}$  &55.8 / 57.6 & 33.3 / 35.7 & \textbf{44.6} / \textbf{46.7}    \\
$Trans_{2} $ & 55.8 / 57.6 & 32.9 / 35.3 &  44.4 / 46.5  \\ \hline

\end{tabular}

\label{table:mapping}
\end{table}

\textbf{Vii) The Structure Analysis of BGAN:} The generator $G$ and discriminator $D$ in BGAN consist of multiple layers of one-dimensional convolution networks. The performance of $G$ and $D$ directly impacts the performance of BGAN. To assess their impact, we conduct experiments using various combinations of one-dimensional convolution layers for $G$ and $D$ based on the Predcls task of Transformer model. The results are presented in \cref{table:STR_WGAN}. Based on the combination (5, 3) of $G$ and $D$ (last row in the table), we individually keep the number of layers fixed for $G$ and $D$ while modifying the number of layers for the other. It is evident that among these combinations, the combination (5, 3) yields the best performance for both $G$ and $D$.

\begin{table}[!htb]
\caption{The structure analysis towards $G$ and $D$ in BGAN.}
\centering
\small
\setlength{\tabcolsep}{1mm}
\begin{tabular}{cc|ccc}
\hline
     \multicolumn{2}{c}{BGAN}    &\multicolumn{3}{|c}{PredCls}    \\  \hline
 $G$(layers) & $D$(layers) &{R@50/100}    & mR@50/100   & M@50/100    \\  \hline
5 & 2 & 55.0 / 56.2 & 33.9 / 36.3  & 44.5 / 46.3   \\
5 & 4 &  60.4 / 62.1 & 28.7 / 31.0  & 44.6 / \textbf{46.7}     \\ \hline
4 & 3 & 57.0 / 59.1  & 32.2 / 34.1   & \textbf{44.6} / 46.6   \\
6 & 3 & 56.9 / 58.7  & 32.1 / 34.3 & 44.5 / 46.5   \\ \hline
5 & 3  &55.8 / 57.6 & 33.3 / 35.7 & \textbf{44.6} / \textbf{46.7}  \\ \hline
\end{tabular}

\label{table:STR_WGAN}
\end{table}

\textbf{Viii) The Effect of Small Non-Zero Value $\varepsilon$:} In constructing the correction bias set (Section \textcolor{red}{3.2}), we utilize the $\varepsilon$ which is set to 0.0001.  In order to assess the impact of $\varepsilon$ for SBG, we conduct the Predcls task using the Transformer model. We test a range of values (0.001, 0.0001, 0.00001) for $\varepsilon$, and the performance of our SBG is presented in \cref{table:ll}. It can be observed that the M@50/100 metric achieves the highest performance when $\varepsilon$ is set to 0.0001, indicating optimal comprehensive performance.

\begin{table}[!htb]
\caption{The effect of small non-zero value $\varepsilon$.}
 \small
\centering
\setlength{\tabcolsep}{1mm}
\begin{tabular}{c|ccc}
\hline
        &\multicolumn{3}{c}{PredCls}    \\  \hline
  $\varepsilon$ value & {R@50/100}    & mR@50/100   & M@50/100    \\  \hline
0.001 & 56.6 / 58.4 & 32.5 / 34.8 & \textbf{44.6} / 46.6   \\
0.0001 &55.8 / 57.6 & 33.3 / 35.7 & \textbf{44.6} / \textbf{46.7}    \\
0.00001 & 56.5 / 58.3 & 32.6 / 34.9 &  \textbf{44.6} / 46.6  \\ \hline

\end{tabular}

\label{table:ll}
\end{table}

\begin{figure*}[!ht]

    \begin{center}
        \includegraphics[width=0.6\columnwidth]{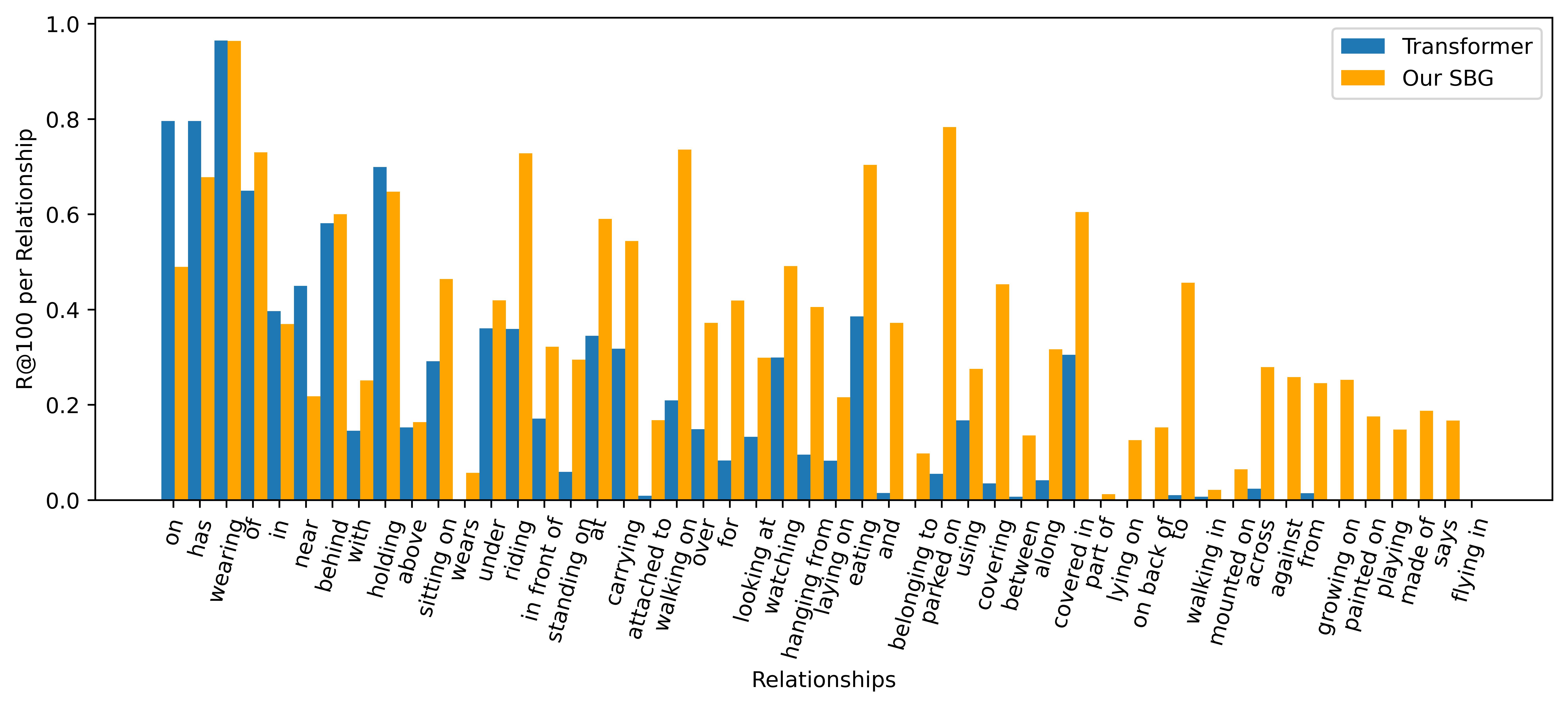}
    \end{center}
  
        \caption{Comparison of R@100 on Transformer and our SBG. The relationships are listed by the long-tailed order. Only ``flying in'' is not improved, whose training samples are only 5, affecting the correction effect of our method.
        }
    \label{fig:R100}

\end{figure*}

\textbf{iX) The Improvements of Long-Tailed Classes:} In Fig.~\ref{fig:R100}, we present the R@100 of each relationship for the PredCls task, comparing  Transformer and our SBG. It shows that all tail classes are improved significantly.

\begin{figure*}[!ht]

    \begin{center}
        \includegraphics[width=0.8\columnwidth]{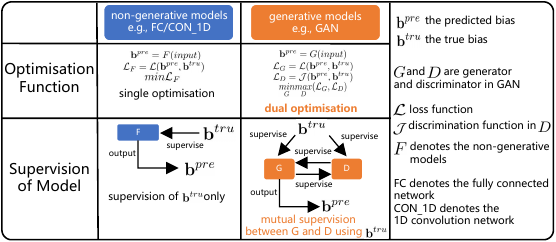}
    \end{center}
 
    \caption{Comparison of generative and non-generative models. }
    
    \label{fig:GAN}
\end{figure*}

\textbf{X) The Rationale for Generative Model.} The bias in our SBG is non-linear and its continuity is very important for correction, so we compare generative models with non-generative models for bias prediction in Fig.~\ref{fig:GAN}. GAN has the dual optimisation that helps to predict the more non-linear bias, and that $G$ and $D$ of GAN supervise each other and promote each other making the $\mathbf{b}^{pre}$ predicted by GAN more closely approximate to the $\mathbf{b}^{tru}$ and capture the continuity of the $\mathbf{b}^{tru}$ better. These are also reflected in HiFi-GAN~\cite{kong2020hifi} and VCA-GAN~\cite{kim2021lip}.

\section{Visualization for Bias Correction}
\begin{figure*}[!htp]
    \begin{center}
        \includegraphics[width=\columnwidth]{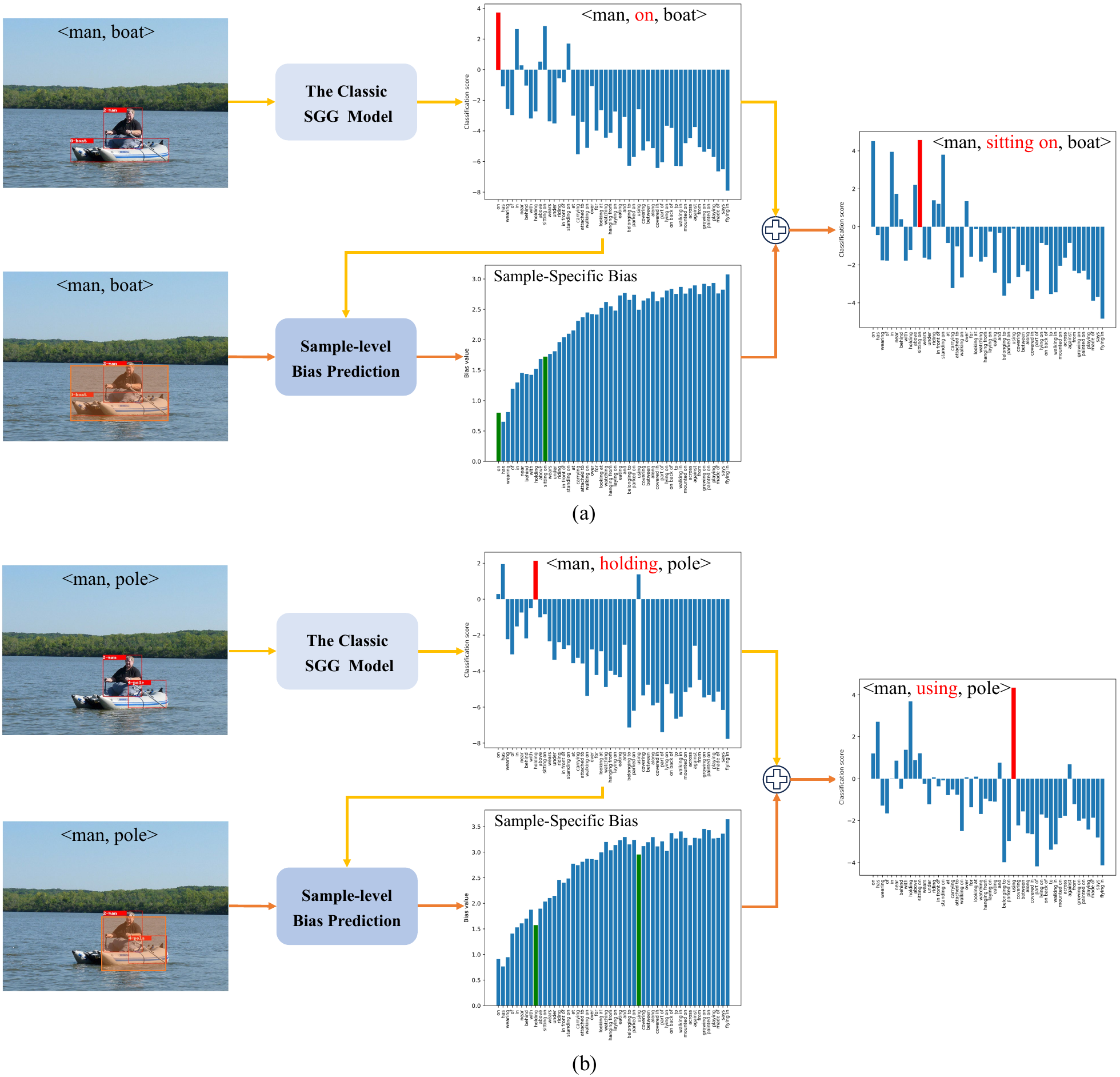}
    \end{center}
   
    \caption{The bias corrections of relationships for object pairs $<$man, boat$>$ and $<$man, pole$>$.}
    \label{fig:vis_add}
\end{figure*}

In order to specifically demonstrate the process of sample-level bias correction, we illustrate the corrections of relationships for object pairs $<$man, boat$>$ and $<$man, pole$>$ as depicted in \cref{fig:vis_add} (a) and \cref{fig:vis_add} (b). The original predictions are the coarse-grained relationships of ``\emph{on}'' and ``\emph{holding}''. Utilizing the contextual information (from union region) of $<$man, boat$>$ and $<$man, pole$>$, the relationships' global bias, and the original predictions, the generator in BGAN predicts the sample-specific biases to refine the coarse-grained ``\emph{on}'' and  ``\emph{holding}'' to the fine-grained ``\emph{sitting on}'' and ``\emph{using}''.

\end{document}